%% file: main.tex
\documentclass[letterpaper, 10 pt, conference]{ieeeconf}  

\IEEEoverridecommandlockouts                              

\usepackage{amsmath}                
\usepackage{amssymb}                
\usepackage[dvipsnames]{xcolor}                 
\usepackage{bm}         
\usepackage{graphicx}
\usepackage{booktabs}
\usepackage{algorithm}
\usepackage{algpseudocode}

\usepackage{comment}

\usepackage{amsthm}

\newtheorem{lemma}{Lemma}
\newtheorem{remark}{Remark}

\newtheorem{proposition}{Proposition}
\newenvironment{proofsketch}%
  {\noindent\hspace{2em}{\itshape Proof sketch: }}%
  {\hspace*{\fill}~\QED\par}
\usepackage{hyperref}
\hypersetup{%
  colorlinks=true,%
  linkcolor={red!50!black},
  citecolor={blue!65!black},
  urlcolor={blue!80!black},
  bookmarksnumbered=true,%
  bookmarksopen=true}
\usepackage[capitalise]{cleveref}
\usepackage{caption}
\makeatletter

\newcommand{\pushright}[1]{\ifmeasuring@#1\else\omit\hfill$\displaystyle{#1}$\fi\ignorespaces}
\newcommand{\pushleft}[1]{\ifmeasuring@#1\else\omit$\displaystyle{#1}$\hfill\fi\ignorespaces}

\let\oldsine\sin
\let\oldcos\cos

\renewcommand{\sin}{\@ifnextchar\bgroup\sin@arg\oldsine}
\renewcommand{\cos}{\@ifnextchar\bgroup\cos@arg\oldcos}

\newcommand{\sin@arg}[1]{s_{#1}}
\newcommand{\cos@arg}[1]{c_{#1}}

\AddToHook{env/algorithmic/before}{\def\@currentcounter{ALG@line}}

\makeatother

\crefalias{ALG@line}{line}   

\title{\LARGE \bf
Graphs of Trajectory Bundles for Nonlinear Motion Planning}

\author{Joris Verhagen$^{1}$, Jon Arrizabalaga$^{2}$, Jana Tumova$^{1}$, and Zachary Manchester$^{2}$
\thanks{$^{1}$Authors are with the Division of Robotics, Perception, and Learning, KTH Royal Institute of Technology, Stockholm, Sweden
        {\tt\small \{jorisv, tumova\}@kth.se}}%
\thanks{$^{2}$Authors are with the Department of Aeronautics and Astronautics, MIT, Cambridge, MA, USA
        {\tt\small \{jonarri, zacm\}@mit.edu}}%
}

\begin{document}

\maketitle

\begin{abstract}

We present the Graphs of Trajectory Bundles (GTB) method for motion planning of systems with nonlinear and black-box dynamics, combining sampling-based motion planning with trajectory optimization and graph search.
Based on a graph-of-convex-sets decomposition of the workspace, GTB iteratively constructs a graph of trajectory bundles that captures the local dynamics of the system at any vertex.
By solving a shortest-path problem on this graph, we find maximally dynamically feasible iterates that satisfy the system's dynamics in the limit and any constraints such as collision avoidance or additional safety or specification constraints by construction of the underlying graph.
As a result, GTB can handle highly nonlinear, unstable, and even black-box dynamics with complex collision and specification constraints, making it a powerful tool for motion planning of complex systems. 

\end{abstract}

\input{1_introduction.tex}

\input{2_preliminaries.tex}

\input{3_algorithm.tex}

\input{4_graph_construction.tex}

\input{5_gtb.tex}

\input{6_results.tex}

\input{7_conclusion.tex}

\bibliographystyle{IEEEtran}
\bibliography{references}


\end{document}

%% file: 1_introduction.tex
\section{Introduction}
We are interested in solving the general motion planning problem of the form
\begin{equation}\label{eq:prob1}
\begin{aligned}
    \min_{x,u} \quad & \sum_{k=0}^{N} J(x_k,u_k) \\
    \text{s.t.} \quad 
    & x_{k+1} = f(x_k,u_k) \\
    & x_k \in \mathcal{W}_{free} \\
    & u_k \in \mathcal{U}, \quad x \in \mathcal{C},
\end{aligned}
\end{equation}
where $\mathcal{W}_{free}$ denotes a (non-convex) collision-free set, $\mathcal{U}$ is a convex actuation set, and $\mathcal{C}$ is an additional constraint set that, e.g., encodes final state, safety constraints, or logic specifications. The system's dynamics,
$f(x_k,u_k):\mathcal{X} \times \mathcal{U} \to \mathcal{X}$, are assumed continuous but otherwise permitted to be nonlinear or \emph{black-box}: required to only be available as a forward map (a numerical integrator, a physics engine, or a learned model) as Jacobians may be unavailable, unreliable, or non-smooth. 
\\\\
The nonconvexities from 1) collision avoidance constraints and 2) nonlinear or black-box dynamics make this problem especially difficult to solve. 
The first challenge can be reasoned over efficiently by considering collision-avoidance as traversal over a graph of convex, collision-free sets, which has been shown to yield globally optimal solutions for differentially flat systems~\cite{marcucci2024shortest}.
While this method is not applicable for systems that are differentially flat with actuation constraints, general nonlinear, or black-box since the dynamic constraints are not convex, its decision-making over sequences of sets via convex optimization is still extremely powerful.
We build upon this idea to reason about the second challenge by collecting sampled trajectories into \emph{bundles}~\cite{tracy2025trajectory} and constructing a \emph{graph of trajectory bundles} over which we can reason about the dynamics in a convex manner via interpolation over the sampled trajectories.
This sampling- and optimization-based approach iteratively refines the trajectory by resampling around the current solution, yielding solutions that satisfy collision-avoidance and specification constraints by construction, and dynamic feasibility in the limit of full algorithm convergence.

Concretely, we formulate \cref{eq:prob1} as a \emph{graph of trajectory bundles} through which we iteratively seek the path of minimal accumulated dynamic infeasibility, i.e., the shortest path in dynamics space from start to goal, while iteratively resampling and refining the dynamics approximation.


\begin{figure}[t!]
    \centering
    \includegraphics[width=0.45\textwidth]{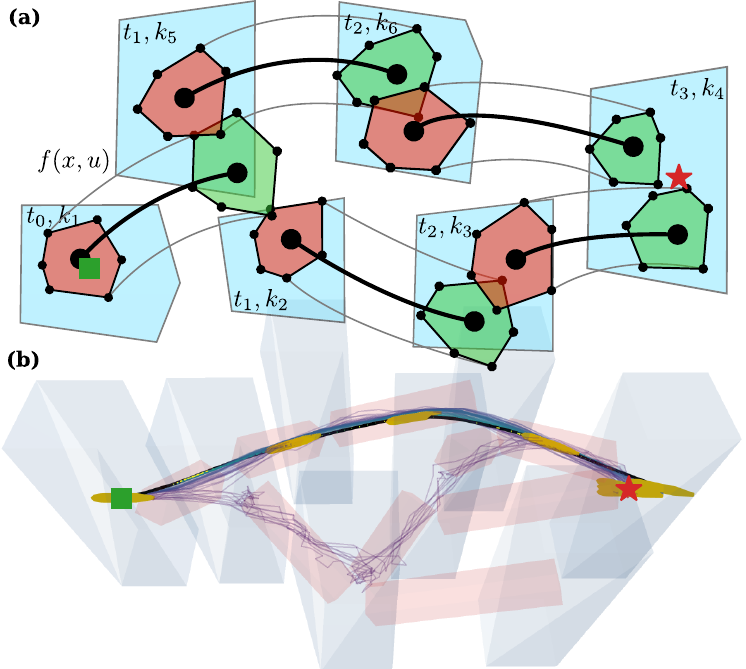}
    \caption{The Graph of Trajectory Bundle (GTB) method poses the motion planning problem as a shortest path problem in dynamics space. On each vertex of a graph of convex sets (shaded blue),  states are sampled (shaded red) and transported through the dynamics $f(x,u)$ to form a bundle (shaded green), themselves being convex sets. 
    At each iteration, GTB finds the interpolators of the sequence of bundles. (b) a high-dimensional underwater robot through a 3D maze with iterations of GTB taking different paths.}
    \label{fig:sample_projection}
    \vspace{-6mm}
\end{figure}

\subsection{Contributions}
This work makes the following specific contributions:
\begin{itemize}
    \item \textbf{Graphs of Trajectory Bundles:} We propose a novel formulation of the problem in \cref{eq:prob1} as a graph-planning problem over sampled trajectory bundles.
    \item \textbf{Sequential Convex Solution Approach:} We solve the Graphs-of-Trajectory-Bundles problem with a sequential convex programming approach on a graph of sampled trajectories.
\end{itemize}

\subsection{Related Work}
The problem in \cref{eq:prob1} is incredibly general and has been studied extensively in many shapes and forms.
Arguably its most fundamental challenge relates to nonconvexities, making it difficult to find optimal or even feasible solutions.

Popular approaches that maintain a global perspective over the solution space include graph-based planning methods such as Probabilistic Roadmaps (PRM)~\cite{kavraki1995random} or A*. 
These methods search for a path over a graph of feasible states. 
By constructing vertices of feasible states and edges of feasible transitions, these methods can find globally optimal solutions, limited by the resolution of the graph.
Recently, the Graph of Convex Sets (GCS) method~\cite{marcucci2024shortest} has been proposed to solve \cref{eq:prob1} for differentially flat dynamics. GCS generalizes discrete graph search to problems where vertices and edges can be convex sets and convex costs, respectively.
Extensions have been proposed to enable non-convex constraints of B\'ezier trajectory optimization~\cite{von2024using} and piecewise affine approximations of nonlinear edge costs~\cite{clark2025plan}, hybrid contact modes for manipulation~\cite{graesdal2024towards}, and forward search for contact-implicit decision-making~\cite{chia2024gcs}. 
Yet none tackle explicit nonlinear or black-box dynamics.

Trajectory optimization methods, such as nonlinear programming~\cite{mir2022survey} and sequential convex programming~\cite{schulman2013finding,schulman2014motion}, have been widely used to find locally optimal solutions to \cref{eq:prob1}. 
These methods rely on gradient-based optimization of a differentiable model; they are sensitive to initial conditions, may converge to local minima, and do not apply when $f$ is non-smooth or available only through simulation.

Derivative-free optimization~\cite{powell2007view,larson2019derivative} and sampling-based trajectory optimization can optimize trajectories without requiring gradient information or explicit dynamics. 
The most popular example in robotics is likely Model Predictive Path Integral control (MPPI)~\cite{williams2017model}. 
These methods naturally deal with nonlinear dynamics and, with advances in simulators and parallelization, can scale to high-dimensional problems~\cite{homburger2025gauss} and multi-modal reasoning~\cite{zhang2024multi}.
Single-shooting methods are, however, unsuitable for long-horizon planning and cannot deal with explicit state constraints~\cite{williams2017model}.
Additionally, for planning in complex geometry, the \emph{tail-wagging-the-dog} problem can lead to infeasible solutions~\cite{tracy2025trajectory}.
The Trajectory Bundle Method (TBM)~\cite{tracy2025trajectory} generalizes MPPI by using a sequence of sample trajectories, referred to as trajectory bundles. This allows constraint satisfaction at every knot point, making it suitable for long-horizon problems with more complex constraints.
However, it is still a local method, requiring a good initial guess to converge to local minima. 




%% file: 2_preliminaries.tex
\section{Preliminaries}
We present Graphs of Trajectory Bundles (GTB), which combines the global reasoning of GCS for obstacle avoidance with the local derivative-free reasoning of the Trajectory Bundle Method (TBM) for dynamic feasibility.

\subsection{Graphs of Convex Sets}
Graph of Convex Sets (GCS)~\cite{marcucci2023motion} generalizes the shortest-path problem by constraining vertices to convex sets and defining convex costs between them.
Let $\mathcal{G}=(\mathcal{V},\mathcal{E})$ be a directed graph in which every vertex $v\in\mathcal{V}$ is equipped with a compact convex set $\mathcal{X}_v$ and a variable $x_v\in\mathcal{X}_v$, and every edge $e=(v,v')\in\mathcal{E}$ a convex cost $\ell_e(x_{v},x_{v'})$.
The shortest path problem seeks both a path $p$ from a source $x_s$ to a target $x_t$ and the points along it,
\begin{equation}\label{eq:spp-gcs}
\begin{aligned}
    \min_{p,\,\{x_v\}} \quad
        & \sum_{e=(v,v')\in\mathcal{E}_p} \ell_e(x_v,x_{v'}) \\
    \text{s.t.} \quad & p\in\mathcal{P}, \\
                      & x_v\in\mathcal{X}_v, \quad \forall v\in\mathcal{V} ,
\end{aligned}
\end{equation}
where $\mathcal{P}$ is the set of source-to-target paths and $\mathcal{E}_p$ the edges traversed by $p$.
Fixing $x_v$ makes \eqref{eq:spp-gcs} a standard shortest path problem with a known Linear Programming (LP) formulation, while fixing the path $p$ makes it a convex program in $x_v$.
\Cref{eq:gcs-flow} below is a concrete formulation of \eqref{eq:spp-gcs} in which the path variable $p$ is replaced by a binary flow $\phi_e\in\{0,1\}$ on every edge,
\begin{equation}\label{eq:gcs-flow}
\begin{aligned}
    \min_{\phi,\,x} \quad & \sum_{e\in\mathcal{E}} \phi_e\,\ell_e(x_{v},x_{v'}) \\
    \text{s.t.} \quad
    & \textstyle\sum_{e\in\mathcal{E}_s^\text{out}}\phi_e
      = \sum_{e\in\mathcal{E}_t^\text{in}}\phi_e = 1, \\
    & \textstyle\sum_{e\in\mathcal{E}_v^\text{in}}\phi_e
      = \sum_{e\in\mathcal{E}_v^\text{out}}\phi_e \leq 1,
      \quad \forall v\in\mathcal{V}\setminus\{s,t\}, \\
    & \phi_e \in \{0,1\}, \quad x_v\in\mathcal{X}_v, \quad \forall v \in \mathcal{V}_p,
\end{aligned}
\end{equation}
where the flow conservation and degree constraints confine $\phi$ to the $s$-$t$ path polytope.
The bilinear product $\phi_e\,\ell_e(x_u,x_v)$ can be written in a relaxed convex form via a binary relaxation, $\phi_e \in [0,1]$, and a perspective transformation. 
We refer to~\cite{marcucci2024shortest} for details.
The GCS formulation is, empirically, a tight relaxation meaning that \eqref{eq:gcs-flow} can be solved efficiently via a single second-order cone program (SOCP).

GCS is applied to motion planning by letting $\mathcal{G}$ be a coverage of the collision-free space with convex regions $\mathcal{Q}_k$, one per vertex, and letting $\mathcal{X}_v$ hold the control points of a B\'ezier curve constrained to $\mathcal{Q}_k$~\cite{marcucci2023motion}.
This requires the trajectory to be an affine function of the vertex variables: the curve is planned in a differentially flat output and dynamic feasibility is assumed to follow from flatness. This assumption means that nonlinear or black-box dynamics of \eqref{eq:prob1} are excluded; moreover, even for a flat model, actuation limits and state-dependent couplings can leave the planned flat-output path infeasible for the real system.

\subsection{Function approximation by interpolation}
To enable (sequential) convex motion planning on the aforementioned nonlinear and black-box systems requires a convex approximation of the dynamics and constraint functions.
An affine approximation can be obtained by Taylor expansion, but requires Jacobians, which may be expensive, unreliable, or unavailable.
An alternative is derivative-free approximation by interpolating sampled function values~\cite{powell1994direct}.
Following the notation in~\cite{tracy2025trajectory}, a weight vector in the simplex $\alpha \in \Delta^{M_s}=\{\alpha \in \mathbb{R}^{M_s}_{\geq0},\ \mathbf{1}^\top\alpha=1\}$ can interpolate over $M_s$ samples of a function $p:\mathbb{R}^{n_y}\to\mathbb{R}^{n_p}$.
For any $p$, we can form an affine approximation around $\bar{y}$ by drawing samples around $\bar{y}$ and placing it in a data matrix
\begin{equation*}\label{eq:interp-matrices}
    W_y = \bigl[y^{(1)}\ \cdots\ y^{(M_s)}\bigr], \quad
    W_p = \bigl[p(y^{(1)})\ \cdots\ p(y^{(M_s)})\bigr],
\end{equation*}
where the affine approximation follows
\begin{equation}\label{eq:interp-approx}
    y = W_y\,\alpha, \qquad p(W_y\alpha) \;\approx\; W_p\,\alpha.
\end{equation}
This derivative-free approximation is exact at every sample and an affine approximation in between.

\subsection{The Trajectory Bundle Method}
\begin{figure*}[t!]
    \centering
    \includegraphics[width=\textwidth]{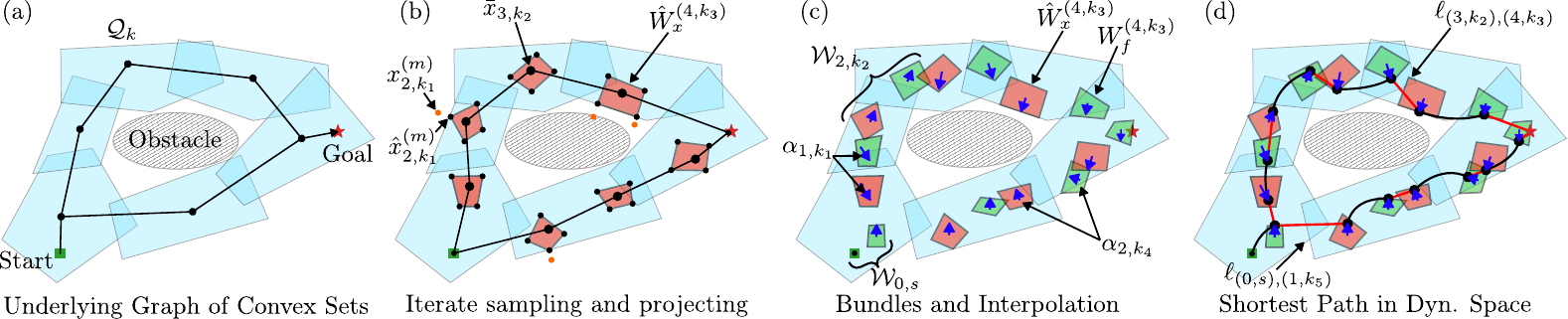}
    \caption{The workspace graph of convex sets (a) and one iteration of the Graphs of Trajectory Bundles (GTB) algorithm (b,c,d). In (b) we perturb the current vertex solution $\bar{x}_{i,k}$ (red samples) and project onto the region $\mathcal{Q}_k$ (blue) to form the \emph{projected bundle} $\hat{W}_x^{(i,k)}$. In (c) each projected sample is rolled through the dynamics to form the \emph{forward bundle} $W_f^{(i,k)}$ (green); the two together are the bundle $\mathcal{W}_{i,k}$ at vertex $(i,k)$ of $\hat{\mathcal{G}}_{\sigma}$. The vertex variables are the weights $\alpha_{i,k}$, chosen in (d) to minimize dynamic infeasibility (red lines) along a path (black curves) from Source to Target~\eqref{eq:gcs_sampled}. 
    }
    \label{fig:algorithm}
    \vspace{-3mm}
\end{figure*}

The Trajectory Bundle Method (TBM)~\cite{tracy2025trajectory} applies this construction to trajectory optimization.
The trajectory is discretized at $N$ \emph{knots}, $i=0,\ldots,N-1$.
Suppose we draw $M_s$ state-control samples $(x_i^{(m)},u_i^{(m)})$ at every knot $i$. We roll out the dynamics, $f_i^{(m)} = f(x_i^{(m)},u_i^{(m)})$, and evaluate a residual (e.g. actuation cost and tracking error), $r_i^{(m)}=r(x_i^{(m)},u_i^{(m)})$, and constraint (e.g. collision avoidance), $c_i^{(m)}=c(x_i^{(m)},u_i^{(m)})$, collecting the data matrices
\begin{equation*}\label{eq:tbm-bundles}
\begin{aligned}
    W_x^{(i)} &= \bigl[x_i^{(1)}\ \cdots\ x_i^{(M_s)}\bigr], &
    W_u^{(i)} &= \bigl[u_i^{(1)}\ \cdots\ u_i^{(M_s)}\bigr], \\
    W_f^{(i)} &= \bigl[f_i^{(1)}\ \cdots\ f_i^{(M_s)}\bigr], &
    W_r^{(i)} &= \bigl[r_i^{(1)}\ \cdots\ r_i^{(M_s)}\bigr], \\
    W_c^{(i)} &= \bigl[c_i^{(1)}\ \cdots\ c_i^{(M_s)}\bigr].
\end{aligned}
\end{equation*}
The pair $\bigl(W_x^{(i)},W_f^{(i)}\bigr)$ is a \emph{bundle} at knot $i$: a sampled, local approximation of the dynamics (the green-red pair in \cref{fig:sample_projection}; formed per region and knot in \cref{fig:algorithm}c).
If the weight vector $\alpha_i$ is shared by all data matrices of knot $i$, trajectory optimization can be approximated by the convex program
\begin{align}\label{eq:tbm-qp}
    \min_{\boldsymbol{\alpha},\,\boldsymbol{s},\,\boldsymbol{w}} \quad
    & \sum_{i=0}^{N-1} \bigl\|W_r^{(i)}\alpha_i\bigr\|_2^2
      + \mu \sum_{i=0}^{N-2} \|s_i\|_1
      + \mu \sum_{i=0}^{N-1} \|w_i\|_1 \\
    \text{s.t.}\quad
    & W_x^{(i+1)}\alpha_{i+1} = W_f^{(i)}\alpha_i + s_i, \notag\\
    & W_c^{(i)}\alpha_i + w_i \geq 0, \notag\\
    & \alpha_i \in \Delta^{M_s}, \qquad \forall i, \notag
\end{align}
where $\mu \in \mathbb{R}_{>0}$ is a penalty weight.
This is a convex problem and an affine approximation of the original problem and returns a trajectory that minimizes the dynamics residual, $\|s_i\|_1$, and constraint violation, $\|w_i\|_1$, while minimizing the cost.
Solving \cref{eq:tbm-qp}, resampling around the new trajectory and re-solving gives a derivative-free SCP method that converges to a local minimum of the original problem~\cite{tracy2025trajectory}.

%% file: 3_algorithm.tex
\section{The Graphs of Trajectory Bundles Algorithm}
\label{sec:sequential-gob}

We solve \eqref{eq:prob1} with a sequential, sampling-based scheme that we call the \emph{Graphs of Trajectory Bundles} (GTB) method, summarized in \cref{alg:gtb} and depicted in \cref{fig:algorithm}.
We start with a \emph{workspace} graph of convex sets, $\mathcal{G}$, whose vertices are feasible convex regions at each time knot.
On this \emph{workspace} graph, we construct, iteratively, a graph of sampled dynamics which, for each vertex in $\mathcal{G}$, approximate the dynamics with an implicit trust region.
In other words, each vertex in $\mathcal{G}$ is replaced by a \emph{projected bundle}: the convex hull of samples projected onto that region and propagated through the dynamics (\cref{sec:bundles-as-sets}). 
GTB therefore solves a sampled, inner approximation of the workspace GCS that considers the dynamics and is refined over the iterations of the algorithm via a contracting process. 
Each iteration $i$ executes four steps:

\begin{enumerate}
    \item \textbf{Sample \& project} — perturb the current per-vertex solution at every knot and every region and project the samples onto that region, \cref{fig:algorithm}b.
    \item \textbf{Build bundles} — propagate every sample through the dynamics to obtain forward states, durations and cost residuals, \cref{fig:algorithm}c, \cref{line:bundle}.
    \item \textbf{GTB region selection \& trajectory} — solve a Graph of Convex Sets problem whose vertices are projected bundles, finding a shortest path with respect to \emph{dynamic infeasibility} by interpolating between the samples in each bundle, \cref{fig:algorithm}d, \cref{line:solve}.
    \item \textbf{Fractional flow updates} — update the vertex solutions for subsequent iterations and compute edge- and vertex probabilities for graph cutting, \cref{line:vertex_sols}.
\end{enumerate}


\begin{algorithm}[t]
\caption{Graphs of Trajectory Bundles (GTB)}
\label{alg:gtb}
\begin{algorithmic}[1]
\Require workspace graph $\mathcal{G}$, source $x_s$, target $x_t$
\State Initialize sampling scale $\sigma$
\State Set initial vertex solutions $(\bar{x}_{i,k},\bar{u}_{i,k})$
\Repeat 
    \State Build projected bundles on each vertex, $\mathcal{W}_{i,k}$ \label{line:bundle}
    \State Construct sampled graph $\hat{\mathcal{G}}_{\sigma}$, solve relaxation \label{line:solve}
    \State Obtain vertex probabilities from $(\phi_e,\alpha_e), \forall e\in\mathcal{E}$ \label{line:probs}
    \State Overwrite vertex solutions $(\bar{x}_{i,k},\bar{u}_{i,k})$ and $\sigma$ \label{line:vertex_sols}
\Until{dynamic infeasibility $\le \texttt{tol}$}
\State \Return path $p$ and trajectory $(x,u)$ rounded from $\phi$
\end{algorithmic}
\end{algorithm}

The sample scaling is annealed across iterations, so that the bundles contract around the persistent vertex solutions; this annealing improves the local estimates of the dynamics and drives the relaxation towards tightness.

%% file: 4_graph_construction.tex
\section{Trajectory Bundles as sets in a graph}
\label{sec:bundles-as-sets}
We now describe the \emph{workspace} graph and the sampled inner approximation used in GTB.

\subsection{Directed Acyclic Graph of Convex Sets}
\label{sec:graph-structure}
Let $\{\mathcal{Q}_k\}_{k=1}^{K}$ be connected convex regions that sufficiently covers $\mathcal{W}_\text{free}$. 
We build a layered Directed Acyclic Graph (DAG) $\mathcal{G}=(\mathcal{V},\mathcal{E})$ with
\begin{equation*}\label{eq:graph-vertices}
    \mathcal{V} = \bigl\{(i,k)\;:\; i\in\{0,...,N-1\},\; k\in\{1,...,K\}\bigr\}
                  \;\cup\; \{s,\,t\},
\end{equation*}
i.e. one \emph{layer} per knot $i$ and one vertex per region $k$ within each layer, plus a source $s$ and a target $t$. 
Edges only connect consecutive layers,
\begin{equation*}\label{eq:graph-edges}
    \begin{aligned}
        \mathcal{E}_\text{int} &= \bigl\{\,(i,j)\!\to\!(i{+}1,k)\;:\;
            j \cap k \neq \emptyset \,\bigr\}, \\
        \mathcal{E}_\text{src} &= \{\,s\!\to\!(0,k):k\in\mathcal{S}_s\,\}, \\
        \mathcal{E}_\text{tgt} &= \{\,(N{-}1,k)\!\to\!t:k\in\mathcal{S}_t\,\}, \\
        \mathcal{E} &= \mathcal{E}_\text{int} \cup \mathcal{E}_\text{src} \cup \mathcal{E}_\text{tgt},
    \end{aligned}
\end{equation*}
where $j\cap k \neq \emptyset$ denotes intersection of region $j$ and $k$, permitting self-transitions ($j=k$).
The source connects to the source regions $\mathcal{S}_s=\{k:x_s\in\mathcal{Q}_k\}$ at layer $0$ and the target regions $\mathcal{S}_t=\{k:x_t\in\mathcal{Q}_k\}$ at layer $N{-}1$ connect to the target, so that any $s\!\to\!t$ trace is a valid region sequence $(k_0,\dots,k_{N-1})$ from source to target. 
In GTB, the graph topology is fixed but the convex sets themselves can change; each iteration creates new bundles, providing a new (improved) approximation of the dynamics.

\subsection{Vertex solutions}
\label{ssec:vertex-solutions}
We let every core vertex $(i,k)$ in $\mathcal{G}$ carry a persistent \emph{vertex solution} $(\bar{x}_{i,k}, \bar{u}_{i,k})$ (\cref{fig:algorithm}b): the state and control the planner currently expects at knot $i$ \emph{if} the trajectory passes through region $k$. 
At the first iteration there is no trajectory information, so they are initialized to the center of $\mathcal{Q}_k$ and $\mathcal{U}$.
Every later iteration overwrites $(\bar{x}_{i,k},\bar{u}_{i,k})$ with the flow-weighted bundle point $(\hat{W}_x^{(i,k)}\alpha_{i,k},\,\hat{W}_u^{(i,k)}\alpha_{i,k})$ (\cref{sec:gcs-relaxation}).
It is around these vertex solutions that the sampling procedure creates a sampled inner approximation of $\mathcal{G}$, creating a new graph $\hat{\mathcal{G}}_\sigma$ whose vertices are the projected bundles.

\subsection{Bundle Construction}
\label{ssec:bundle-construction}
At each vertex $(i,k)$ we draw $M_s$ perturbation samples around the vertex solution,
\begin{equation}\label{eq:sample-update}
    x_{i,k}^{(m)} = \bar{x}_{i,k} + \varepsilon_{i,k}^{(m)},\quad
    \varepsilon_{i,k}{(m)}\sim\mathcal{S}(\sigma),\quad
    m=1,\dots,M_s,
\end{equation}
and analogously sample controls $u_{i,k}^{(m)}$ around $\bar{u}_{i,k}$. 
Here, $\mathcal{S}(\sigma)$ denotes a perturbation scheme with scale $\sigma$. 
This includes, e.g., isotropic Gaussian sampling $\mathcal{S}(\sigma)=\mathcal{N}(0,\sigma^2I)$, uniform sampling on the hypercube $\mathcal{S}(\sigma)=\mathcal{U}([-\sigma,\sigma]^{n_x})$, or deterministic (pointwise) perturbations on a grid of radius $\sigma$.
These samples are then projected onto the corresponding convex set, creating projected bundles (\cref{fig:algorithm}b);
\begin{gather*}
    \hat{x}_{i,k}^{(m)} = \Pi_{\mathcal{Q}_k}\bigl(x_{i,k}^{(m)}\bigr), \quad
    \hat{u}_{i,k}^{(m)} = \Pi_{\mathcal{U}}\bigl(u_{i,k}^{(m)}\bigr), \\[2pt]
    \begin{aligned}
        \hat{W}_x^{(i,k)} &= \bigl[\hat{x}_{i,k}^{(1)}\ \cdots\ \hat{x}_{i,k}^{(M_s)}\bigr]
            \in\mathbb{R}^{n_x\times M_s}, \\
        \hat{W}_u^{(i,k)} &= \bigl[\hat{u}_{i,k}^{(1)}\ \cdots\ \hat{u}_{i,k}^{(M_s)}\bigr]
            \in\mathbb{R}^{n_u\times M_s}.
    \end{aligned}
\end{gather*}
Because each $\hat{x}_{i,k}^{(m)}\in\mathcal{Q}_k$ by construction, the bundle requires \emph{no} additional region-membership constraints downstream (\cref{sec:bundles-as-sets}).


Each projected state/control sample $(\hat{x}_{i,k}^{(m)}, \hat{u}_{i,k}^{(m)})$ is propagated through the dynamics with duration $t_{i,k}^{(m)}$ (via e.g. integrating known dynamics or inferring a simulator or black-box dynamics model). 
This leads to a forward state $f_{i,k}^{(m)}=f(\hat{x}_{i,k}^{(m)},\hat{u}_{i,k}^{(m)},t_{i,k}^{(m)})$, all of which are collected into a forward bundle and a duration bundle .
\begin{align*}
    W_f^{(i,k)}&=\bigl[f_{i,k}^{(1)}\ \cdots\ f_{i,k}^{(M_s)}\bigr] \in \mathbb{R}^{n_x\times M_s}\\ 
    W_t^{(i,k)}&=\bigl[t_{i,k}^{(1)}\ \cdots\ t_{i,k}^{(M_s)}\bigr] \in \mathbb{R}_{\geq 0}^{1\times M_s}
\end{align*}
together with residual bundles~\cite{tracy2025trajectory}.
Bundles now live at vertices $(i,k)$ of the graph rather than at knots as in~\cite{tracy2025trajectory}.
Collision avoidance at knot $i{+}1$ is guaranteed by the projected bundle $\hat{W}_x^{(i+1,k)}$, not by $W_f^{(i,k)}$.
The vertex bundles represent the local, sampled approximation of $f(x,u)$ at $(i,k)$ around which the convex subproblem is built.

\subsection{A Graph of Trajectory Bundles}
The workspace graph $\mathcal{G}$ of \cref{sec:graph-structure} is a graph of convex sets in the classical sense: each vertex carries a convex region $\mathcal{Q}_k$. Solving a GCS over those regions directly would require a representation of the nonlinear dynamics valid over all of $\mathcal{Q}_k$, which we do not have. 
Instead we solve a GCS over the \emph{projected bundles}.
First, we stack the projected state, control, duration, cost,and forward state of every sample into a \emph{lifted generator} $g_{i,k}^{(m)} \in \mathbb{R}^{2n_x+n_u+2}$,
\begin{equation}\label{eq:generator}
    g_{i,k}^{(m)} =
    \bigl[\,\hat{x}_{i,k}^{(m)\top}\;\;
            \hat{u}_{i,k}^{(m)\top}\;\;
            t_{i,k}^{(m)}\;\;
            r_{i,k}^{(m)}\;\;
            f_{i,k}^{(m)\top}\,\bigr]^\top.
\end{equation}
We then define a V-polytope from these lifted generators with vertices $V_{i,k}=\bigl[g_{i,k}^{(1)}\ \cdots\ g_{i,k}^{(M_s)}\bigr]$ yielding $\mathcal{W}_{i,k}$ which is the convex hull of \emph{lifted generators} in which any interior point is implicitly a convex combination of the generators, i.e. a weight vector $\alpha_{i,k}\in\Delta^{M_s-1}$ exists implicitly,
\begin{equation}\label{eq:vertex-set}
    \mathcal{W}_{i,k} = \operatorname{conv}\bigl\{g_{i,k}^{(m)}\bigr\}_{m=1}^{M_s}
    = \bigl\{\,V_{i,k}\,\alpha_{i,k}\;:\;\alpha_{i,k}\in\Delta^{M_s-1}\,\bigr\}.
\end{equation}
Although $f$ is nonlinear, each generator $g_{i,k}^{(m)}$ is fixed data so $V_{i,k}\alpha_{i,k}$ is affine in $\alpha_{i,k}$ and $\mathcal{W}_{i,k}$ is a polytope. 
The nonlinearity is absorbed at sampling time, leaving the vertex set, and every edge cost built from it, convex.
This convex set represents an inner, sampled, approximation of the dynamics in $\mathcal{Q}_k$ at step $i$.
A weight vector $\alpha_{i,k}$ fixes a state $x$, control input $u$, propagation $f$, duration $t$, and cost $r$ through the bundle matrices, \eqref{eq:interp-approx}.
To construct the iteration's graph, we replace each region $\mathcal{Q}_k$ at knot $i$ by its projected bundle $\mathcal{W}_{i,k}$, keeping the edge set $\mathcal{E}$ of \cref{sec:graph-structure} unchanged: $\hat{\mathcal{G}}_{\sigma} = (\mathcal{W}_{(i,k)},\mathcal{E})$. 
Recall that $\mathcal{E}$ contains only forward, single-step edges between consecutive time layers, and only between regions that overlap ($j\cap k\neq\emptyset$); an edge is thus a candidate one-step transition from region $j$ at knot $i$ to a spatially adjacent region $k$ at knot $i{+}1$. 


%% file: 5_gtb.tex
\begin{figure*}[t!]
    \centering
    \includegraphics[width=\textwidth]{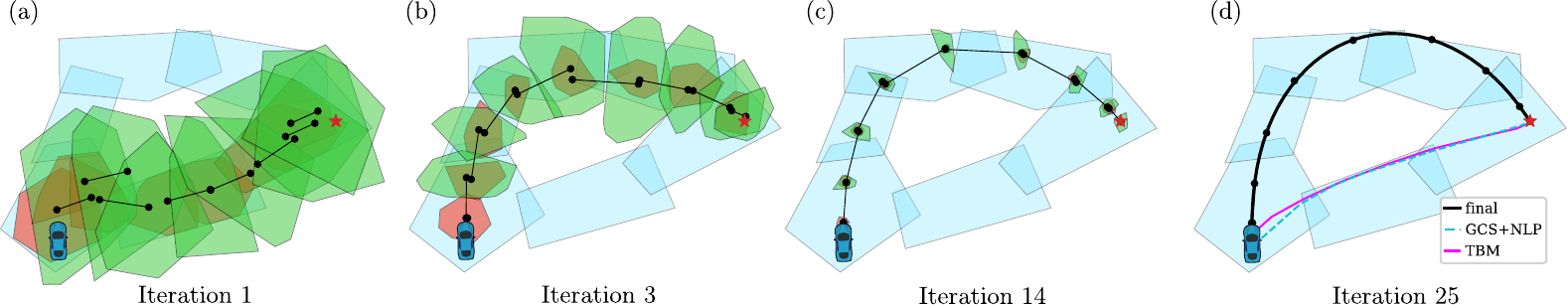}
    \caption{Iterations of GTB for a path planning problem for a car model navigating around an obstacle, showing the projected bundles (shaded red) and forward bundles (shaded green) of the highest-probability path. 
    Early iterations explore modes via the coarse approximation of the dynamics while later iterations focus on refining the trajectory
    (d) shows the final trajectory and a comparison to (dynamically infeasible) results of GCS initializing an NLP solver, which gets stuck on the shortest-path local minima and the Trajectory Bundle Method which gets stuck on the local minima of the initial guess.}
    \label{fig:car_iterations}
    \vspace{-3mm}
\end{figure*}

\section{The GTB Relaxation}
\label{sec:gcs-relaxation}
We now describe how GTB formulates iterative shortest path problems in dynamics space via $\hat{\mathcal{G}}_{\sigma}$ and how, analogously to \cite{marcucci2024shortest}, it can be relaxed to a convex program. 
We then show that the iterative nature of GTB, which contracts the bundles around the vertex solutions, drives the relaxation towards tightness.

\subsection{The dynamic-infeasibility edge cost}
For an interior edge $e=\bigl((i,j)\!\to\!(i{+}1,k)\bigr)\in\mathcal{E}_\text{int}$ we assign the convex \emph{consensus} (dynamic-infeasibility) cost between the forward block of the source vertex and the state block of the destination vertex,
\begin{equation}\label{eq:edge-cost}
    \ell_e\bigl(\alpha_{i,j},\alpha_{i+1,k}\bigr) =
        \bigl\| W_f^{(i,j)}\alpha_{i,j}
              - \hat{W}_x^{(i+1,k)}\alpha_{i+1,k}\bigr\|,
\end{equation}
The boundary edges carry the matching endpoint costs
\begin{equation}\label{eq:boundary-cost}
    \begin{aligned}
        \ell_{s\to(0,k)} &= \bigl\|\hat{W}_x^{(0,k)}\alpha_{0,k}-x_s\bigr\|,\\
        \ell_{(N-1,k)\to t} &= \bigl\|\hat{W}_x^{(N-1,k)}\alpha_{N-1,k}-x_t\bigr\|.
    \end{aligned}
\end{equation}
A valid trace of the graph is then a sequence of regions from $x_s$ to $x_t$ whose accumulated cost \eqref{eq:edge-cost}--\eqref{eq:boundary-cost} is small: the final state reaches the target and, at every layer, the one-step image of the chosen point in $\mathcal{W}_{i,j}$ coincides with the chosen point in the next vertex $\mathcal{W}_{i+1,k}$ (minimal dynamic infeasibility). 
This is a shortest-path problem in \emph{dynamics space} as opposed to a shortest path in, e.g., Euclidean space.
The ``length'' of a path is therefore not how far the system traveled (as in GCS) but how badly a sequence of interpolators fail to create a dynamically feasible trajectory. 

\subsection{A GTB shortest-path iteration}
Collecting the convex vertex sets \eqref{eq:vertex-set} and the dynamic-infeasibility edge costs \eqref{eq:edge-cost}--\eqref{eq:boundary-cost} into a single shortest path problem over $\hat{\mathcal{G}}_{\sigma}$,
\begin{align}\label{eq:gcs_sampled}
    \min_{\boldsymbol{\alpha},\,\boldsymbol{\phi}} \quad
    & \sum_{e \in \mathcal{E}} \phi_e \Bigl[
        \ell_e\bigl(\alpha_{i,j},\alpha_{i+1,k}\bigr) + \mu \bigl\| h_e\bigl(\alpha_{i,j}\bigr) \bigr\| + \nu \bigl\|\alpha_{i,j}\bigr\| \Bigr] \\
    \text{s.t.}\quad
    & \boldsymbol{\phi}\ \text{on the $s$--$t$ path polytope of \eqref{eq:gcs-flow}}, \notag\\
    & \phi_e\in\{0,1\}, \quad \alpha_{i,k}\in\Delta^{M_s-1}, \quad
      \forall e, \forall i, \forall k \notag
\end{align}
Rather than an abstract path variable $p\in\mathcal{P}$, the region sequence is now a \emph{flow encoding}: a binary edge flow $\phi_e$ held on the $s$--$t$ path polytope by the flow-conservation and degree constraints of \eqref{eq:gcs-flow}.
Here $h_e(\alpha_{i,j})$ is an additional running cost, e.g. $W_r^{(i,j)}\alpha_{i,j}$, and $\bigl\|\alpha_{i,j}\bigr\|$ is a regularization term as $M_s > n_x$;
$(\mu, \nu) \geq 0, (\mu, \nu) \ll 1$ are small weights.
A solution to \eqref{eq:gcs_sampled} selects both the flow (a region sequence) and the interpolators $\alpha_{i,k}$ along it, shown schematically in \cref{fig:algorithm}c.
We solve its perspective-form relaxation, relaxing $\phi_e\in\{0,1\}$ to $\phi_e\in[0,1]$ and rewriting each bilinear $\phi_e\,\ell_e$ through the perspective transform~\cite{marcucci2024shortest}.
\paragraph{Norms}
Different norms can be used in \eqref{eq:gcs_sampled}.
The $\ell_2$ and $\ell_2^2$ norms are originally presented in GCS~\cite{marcucci2023motion,marcucci2024shortest}. The $\ell_1$ norm, used in TBM~\cite{tracy2025trajectory}, is also a valid choice and promotes sparsity. Additionally its perspective relaxation is a (larger) LP rather than an SOCP.
We note in \cref{sec:results} that the $\ell_2$ norm is most robust for iterative GTB solves, while the $\ell_1$ norm is more sensitive to the initial bundle sampling and can lead to suboptimal paths.

\subsection{Fractional Flow Solutions}
The relaxed solution provides, for every vertex carrying flow, the points $\bar{x}_{i,k}=\hat{W}_x^{(i,k)}\alpha_{i,k}$ and $\bar{u}_{i,k}=\hat{W}_u^{(i,k)}\alpha_{i,k}$ (the new vertex solution of \cref{ssec:vertex-solutions}), shown in \cref{fig:algorithm}d.

Convex relaxations of GCS shortest-path problems are often \emph{tight}: the optimal flow $\phi$ is integral and traces a single $s\!\to\!t$ path. 
The relaxation of \eqref{eq:gcs_sampled} is not, particularly in early iterations, where bundles are wide, sets overlap, and no single region sequence dominates.
This is not a failure mode that should be rounded away: the fractional flow is a convex combination of integral flows so \eqref{eq:gcs_sampled} returns a \emph{weighted} set of candidate paths with a vertex solution for each region along those paths.
We exploit this in two ways:

\paragraph{Resampling live vertices}
Any vertex solution $(\bar{x}_{i,k},\bar{u}_{i,k})$ on an edge with $\phi_e>0$ is a candidate for the next iterate, and we therefore resample around this solution to create the next bundle $\mathcal{W}_{i,k}$.
This is the powerful aspect of GTB where, rather than committing to a single path, the fractional flow and GCS construction keeps track of a graph of possible paths.
\cref{fig:car_iterations} shows how early iterations explore different paths while later iterations focus on refining the selected path.

\paragraph{Marginals over regions}
The magnitudes of the fractional flow admit a probabilistic interpretation.
For every edge flow $\phi_e \in [0,1]$, satisfying flow conservation~\eqref{eq:gcs-flow}, the edge flow is a probability mass over transitions,
\begin{equation}\label{eq:edge-distribution}
  q_{i,k\to k'} = \frac{\phi_{i,k\to k'}}{\sum_{k''}\phi_{i,k\to k''}}.
\end{equation}
The marginal probability that a path then visits region $k$ at layer $i$ is obtained from the incoming flow
\begin{align}\label{eq:vertex-propagation}
  p_{i,k} &= \sum_{k':\,(i-1,k'\to k)\in\text{in}(k)} q_{i-1,k'\to k} \cdot p_{i-1,k'}, \\
  p_{0,k} &= \phi_{s\to k}, \notag
\end{align}
and by flow conservation, $\sum_{k} p_{i,k} =1$ at every layer $i$.
We accumulate these marginals over iterations in order to disregard vertices that are consistently unlikely, lowering the computational complexity of subsequent iterations.

\subsection{Tightness, Guarantees, and Limitations}
A critical point for a convex relaxation, which is solved as in \cref{eq:gcs_sampled} for every GTB iteration, is its tightness or integrality gap $\Delta$, measuring the degree to which the optimal value of the GCS relaxation, $p_{CP}$, matches the original (non-convex) problem of the binary GCS formulation, $p_{IP}$, for a specific GTB iteration. 
While we cannot prove convergence of GTB, we can show that the relaxation is tight in the limit: $\Delta(\sigma)=p_{IP}(\sigma) - p_{CP}(\sigma) = 0$ as $\sigma \rightarrow 0$.

\begin{lemma}[Tightness of the GCS relaxation]\label{lem:tightness}
    Assume $f(x,u)$ is continuous, $\mathcal{U}$ is compact and convex, and each region $\mathcal{Q}_k$ is a nonempty closed convex set, making the generators $\{\bar{g}_{i,k}\}$ well-defined.
    Then at $\sigma=0$ the integrality gap vanishes, $\Delta(0)=0$.
\end{lemma}

\begin{proofsketch}
    At $\sigma=0$, the bundles are a single point, so $R_{\sigma}$ degenerates to a min-cost $s\to t$ flow with fixed edge weights, and flow polytopes of layered DAGs are integral.
\end{proofsketch}

\begin{proposition}[Optimality gap shrinking]\label{prop:limit}
    Adopt \cref{lem:tightness}, with edge cost $\ell_e=\|\cdot\|_2$ as in \eqref{eq:edge-cost}--\eqref{eq:boundary-cost}, and assume $f$ is locally Lipschitz and the sampling/projection scheme \eqref{eq:sample-update} is nonexpansive with scale $\sigma$.
    Then the integrality gap $\Delta(\sigma)$ is controlled by the bundles' diameter, which itself shrinks linearly with $\sigma$: $\Delta(\sigma)\to0$ continuously as $\sigma\to0$.
    Annealing the bundles down (\cref{sec:sequential-gob}) therefore drives the relaxation smoothly toward the exact, tight limit of \cref{lem:tightness}.
\end{proposition}

\begin{remark}
    The converse of \cref{prop:limit} explains the behavior and the power of GTB solving sequential sampled GCS subproblems. 
    In early iterations, the interpolators show a convex combination of paths as a suitable candidate of the overall solution. 
    Rather than treating this as a failure, GTB interprets it probabilistically, keeping track of promising solution candidates and updating vertex solutions accordingly.
\end{remark}

\begin{remark}[Guarantees and limitations]
    GTB inherits the guarantees of sequentially convex methods. 
    Each iteration solves an inner, sampled approximation of \eqref{eq:prob1}, so the method reaches at best a local minimum; we \emph{cannot} certify global optimality, tightness of the relaxation at a given $\sigma>0$, or that the dynamic-infeasibility cost vanishes as $\sigma\to0$.
    What GTB adds over standard SCP and TBM is a global perspective (the fractional flow keeps several region sequences alive rather than committing to one linearization point) and support for black-box $f$.
    GTB also removes sensitivity to the initial \emph{trajectory} guess, but a weaker dependence on the convex decomposition $\{\mathcal{Q}_k\}$ remains.
\end{remark}



\subsection{Solver Details}

\paragraph{Graph cutting}
The marginals \eqref{eq:edge-distribution}--\eqref{eq:vertex-propagation} are used to accumulate evidence: every edge $e$ carries a persistent probability $P_e$ which is updated after each solve by a log-odds increment of that iteration's marginal.
A region transition has to be \emph{persistently} unused before its probability falls below a threshold $\delta$ and it is dropped from the next relaxation.
Since the number of variables and cones in \eqref{eq:gcs_sampled} scales with $|\mathcal{E}_\text{int}|$, this progressively shrinks the conic program as the run converges, without altering the formulation on the edges that remain.

\paragraph{Annealing the bundle scale}
The bundle scale (and hence the trust region) is annealed via an \emph{exploration} and \emph{exploitation} phase.
First, the bundles are wide enough to explore several region sequences in parallel, and the fractional flow is used to refine the vertex solutions of all live vertices.
When the rounded path settles, the bundles are re-bloated to a trust region wide enough to allow the continuous refinement to still move the trajectory, and then contracted again to polish the solution.

%% file: 6_results.tex
\begin{figure*}[t!]
    \centering
    \includegraphics[width=\textwidth]{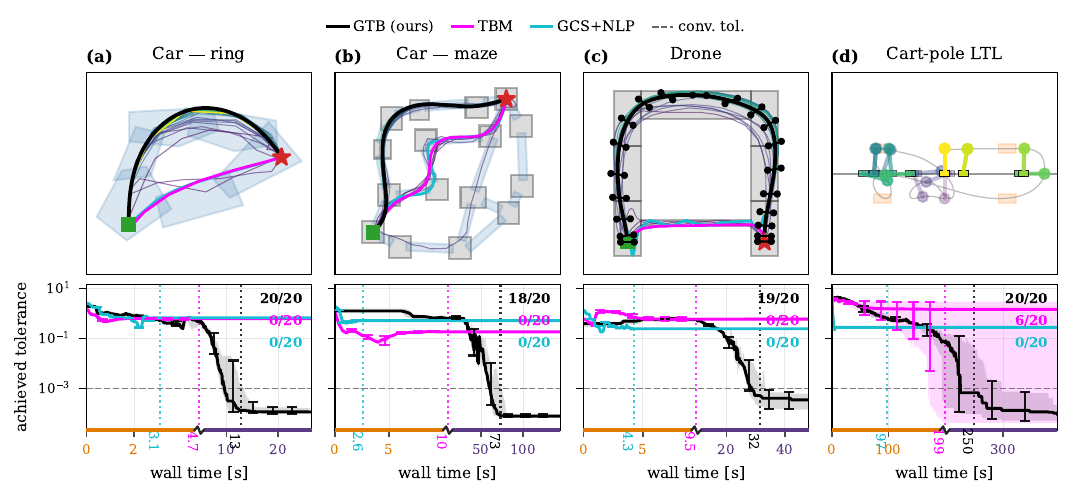}
    \caption{Comparisons of GTB against the Trajectory Bundle Method and the GCS solve with an NLP refinement. 
    On top the physical trajectories and early iterations of the GTB algorithm highlighting the different paths that are considered.
    On the bottom the tolerance as a function of the wall clock time. Dotted vertical lines denote the average computation time. The orange and purple x-axis denote the different scales to highlight that GTB converges approximately equally rapid to infeasible solutions, but converges to lower tolerances afterwards.
    While GTB almost always finds feasible solutions, both TBM and GCS+NLP often fail and are stuck in local minima. 
    }
    \label{fig:final_results}
    \vspace{-5mm}
\end{figure*}

\section{Results}
\label{sec:results}
We demonstrate GTB on motion planning problems for systems with nonlinear or black-box dynamics, satisfying complex collision- and specification constraints.
For all scenarios we sample with a Gaussian distribution and report average results over 20 random seeds.
We solve all scenarios on a desktop with an AMD Ryzen 7 9800X3D CPU and 64 Gb of RAM.
\\\\
We compare GTB solved with Mosek~\cite{aps2019mosek} against the Trajectory Bundle Method (TBM) solved with Clarabel~\cite{goulart2026clarabel} and using GCS as an initial guess for a nonlinear trajectory problem (GCS+NLP) solved with Mosek and IPOPT~\cite{biegler2009large} sequentially. 
In the TBM benchmark, we use the same $\sigma$-annealing and sampling scheme and run the same number of iterations. 
Collision avoidance is enforced via a penalty term in the cost function which is satisfied if any bundle is in at least one convex region. 
For GCS+NLP, the GCS solution is used as an initial guess for a nonlinear solver which enforces collision avoidance for the set that GCS allocated to the segment.
The car and drone models are analytic so that GCS+NLP can be compared against; for the MuJoCo cartpole, which GTB and TBM treat as black-box, GCS+NLP is handed an analytic model.
Details on the scenarios are shown in \cref{tab:hyperparams}.

As both GTB and TBM are iterative sampling-based methods, neither are guaranteed to find a feasible solution and there are always seeds and scenarios where optimal solutions can be found. 
The results highlight that GTB is able to, much more reliably, find feasible solutions on a variety of graph-based motion planning problems, while TBM and GCS+NLP often get stuck in local minima.


\begin{table}[ht]
\centering
\setlength{\tabcolsep}{5pt}
\caption{Per-scenario hyperparameters.}
\label{tab:hyperparams}
\begin{tabular}{lccccc}
\toprule
 & \textbf{Car-ring} & \textbf{Car-maze} & \textbf{Drone} & \textbf{Cart-pole LTL} & \textbf{AUV} \\
\midrule
$N$         & 10             & 30             & 40             & 40             & 40 \\
$M_s$        & 60             & 60             & 100            & 40             & 70 \\
max\_iter             & 100            & 80             & 50             & 200            & 250 \\
$dt$                  & 0.75           & 0.25           & 0.1            & 0.2            & 1.0 \\
$u_{\text{order}}$    & 1              & 1              & 1              & 1              & 2 \\
$n_{\text{regions}}$  & 6              & 35             & 8              & 5              & 17 \\
$\sigma_{\text{start}}$ & 0.75  & 0.75 & 0.75 & 1.0 & 0.5 \\
$\sigma_{\text{end}}$ & 0.005 & 0.01 & 0.002 & 0.002 & 0.001 \\
exploit        & 4.0            & 4.0            & 3.0            & 4.0            & 1.0 \\
\bottomrule
\end{tabular}
\end{table}

\subsection{Car through a maze}
We consider the bicycle model with Ackermann steering, which is a common model for car-like vehicles where heading change is coupled to the forward velocity with $x = [x, y, v, \theta, \delta]^T \in \mathcal{W}_{free}$ and $u = [u_a, u_\delta]^T \in \mathcal{U}$. The dynamics are given by
\begin{equation*}
    \dot{x} =
    \begin{bmatrix}
        \dot{x} \\ \dot{y} \\ \dot{v} \\ \dot{\theta} \\ \dot{\delta}
    \end{bmatrix}
    =
    \begin{bmatrix}
        v \cos(\theta) \\
        v \sin(\theta) \\
        u_a \\
        \frac{v}{L} \tan(\delta) \\
        u_\delta
    \end{bmatrix}
\end{equation*} 
The bicycle model is differentially flat in the position $(x,y)$, so a GCS planner can trace a geometric path through the maze. 
That path, however, ignores input bounds $\mathcal{U}$ on $(u_a,u_\delta)$.
The GTB method in the corridor scenario is shown in \cref{fig:car_iterations} and \cref{fig:final_results}a showing the evolution of the trajectory bundle over iterations and the tolerance over wall-clock time. \cref{fig:final_results}b shows the car through a more complicated maze, substantiating the capability of GTB yet highlighting the additional computation time required to iteratively solve the GCS subproblems.
GTB finds feasible solutions, satisfying the dynamics, input, and the collision constraints in 20 and 18 out of 20 seeds respectively.
For both scenarios, TBM and GCS+NLP fail to find dynamically feasible solutions. The initial guess (straight-line interpolation) or the shortest path formulation respectively bring the system to an infeasible local minimum.


\subsection{Planar drone through a thin gap}
We consider a planar drone with $x = [x, y, \dot{x}, \dot{y}, \theta, \omega]^T \in \mathcal{W}_{free}$ and $u = [u_1, u_2]^T \in \mathcal{U}$ being the thrusts of the two propellors. 
The dynamics are given by
\begin{equation*}
    \dot{x} =
    \begin{bmatrix}
        \dot{x} \\ \dot{y} \\ \ddot{x} \\ \ddot{y} \\ \dot{\theta} \\ \dot{\omega}
    \end{bmatrix}
    =
    \begin{bmatrix}
        \dot{x} \\
        \dot{y} \\
        -\frac{1}{m} (u_1 + u_2) \sin(\theta) \\
        -g + \frac{1}{m} (u_1 + u_2) \cos(\theta) \\
        \omega \\
        -\frac{l}{I} (u_1 - u_2)
    \end{bmatrix}
\end{equation*}
The drone needs a non-zero orientation, $\theta$, to travel horizontally.
The considered scenario has two gaps: a narrow one where collisions can only be avoided if the orientation is severly limited, and a thick one where the drone can orient itself more freely.
GTB reasons over all possible sequences of regions to find one which permits a dynamically feasible trajectory, finding a path through a sequence of convex sets that permits the required orientation in 19 out of 20 seeds.
The results are shown in \cref{fig:final_results}c where GTB finds a feasible solution, satisfying the dynamics and the collision constraints, while both TBM and GCS+NLP fail to find feasible solutions and are again stuck in local (shortest-path) minima.

\subsection{Simulated Cartpole with Temporal Logic}
While GTB permits state-space decomposition to solve maze-like scenarios, we can also use the DAG structure to encode temporal logic specifications, as in~\cite{kurtz2023temporal}.
As the GTB method permits explicit constraints via set-containment, it goes beyond single-shooting methods such as MPPI, encoding temporal logic specifications by construction. 
This makes it possible for unstable nonlinear systems to satisfy temporal logic specifications, which is not possible with TBM or GCS+NLP.
We consider a cartpole defined in Mujoco~\cite{todorov2012mujoco} that, from a downward position, has to satisfy the Linear Temporal Logic (LTL) specification $\phi$, encoding that we either visit \emph{top left} or \emph{top  right} after which we visit both \emph{down left} and \emph{down right};
\begin{equation*}
    \phi = F ( (\text{top left} \vee \text{top right}) \wedge F (\text{down left} \wedge \text{down right}) )
\end{equation*}
and finally end upright in the middle.
\cref{fig:final_results}d shows the GTB method finds a feasible solution, satisfying the dynamics and the temporal logic specification explicitly via the product graph of this unstable black-box system. 
In contrast, TBM and GCS+NLP fail to consistently find feasible solutions. 
While the GCS+NLP method finds a satisfying solution for the specification, but the shortest path formulation limits segments to infeasible transitions when swing-up needs to be considered.
The average computation time is 250 seconds.

\subsection{Autonomous Underwater Vehicle}
We finally show how GTB can also be applied to more complex and learned dynamics models.
The SAM Autonomous Underwater Vehicle (AUV)~\cite{bhat2020cyber} is a torpedo-like robot often modelled via Fossen's model~\cite{fossen1995nonlinear}. 
A challenge for this model is an accurate representation of the damping matrix, which models complex hydrodynamic effects. We learn this damping matrix via a physics-informed neural network (PINN), inspired by~\cite{backne2025physics}, modelling the damping matrix $D$ as $L^T L$ where $L$ is a 2 layer, 64 neuron model. 
The state of the AUV is the position, orientation, and linear/angular velocities, plus two internal states for the pitch and depth actuation; $N_x=15$. 
The control inputs are the propellors, propeller pitch and yaw, as well as pitch and depth actuation; $N_u=6$. Details can be found in~\cite{bhat2020cyber}.
\cref{fig:sample_projection} shows a collision-free trajectory through a maze-like environment while satisfying the dynamics and actuation constraints, even on a complex, high-dimensional system as SAM.
The total computation time is 544 seconds with an overall success rate of 19/20, dominated by several early full-maze solves to discover the feasible sequence of regions.

%% file: 7_conclusion.tex
\section{Conclusion and Future Work}
We presented the Graph of Trajectory Bundles (GTB) method for graph-based motion planning for nonlinear or black-box systems.
GTB relies on constructing a sampled inner approximation of a workspace GCS that considers the dynamics via convex hulls of sampled trajectories. 
Minimizing the dynamics residual can then be formulated as a shortest path problem over the graph which is solved iteratively, converging to feasible trajectories in the limit.

Future work is on improving the solution times of GTB. 
The sequential/iterative nature of the algorithm, where large but sparse graph-structured second-order cone programs need to be solved, could lend itself to custom solvers that utilize e.g. distributed consensus, warm-starting, or tailored factorization methods.
Tailored solvers can then open the door to the line-graph construction of $\hat{\mathcal{G}}$,  richer specifications, more complex environments, and hardware implementation.
